\pdfoutput=1

\documentclass[10pt,twocolumn,letterpaper]{article}

\usepackage{cvpr}
\usepackage{times}
\usepackage{epsfig}
\usepackage{graphicx}
\usepackage{amsmath}
\usepackage{amssymb}
\usepackage{comment}
\usepackage{bm}
\usepackage{subfigure}
\usepackage{xcolor}
\usepackage{ellipsis}
\usepackage{setspace}
\usepackage{nth}

\DeclareMathOperator{\EX}{\mathbb{E}} 
\DeclareMathOperator{\eps}{\epsilon} 

\newcommand{\ttscale}{s*[0.8]}
\DeclareFontFamily{OT1}{cmtt}{\hyphenchar\font -1}
\DeclareFontShape{OT1}{cmtt}{m}{n}{
        <-9>   \ttscale cmtt8
        <9-10> \ttscale cmtt9
        <10-12>\ttscale cmtt10
        <12->  \ttscale cmtt12
}{}
\DeclareFontShape{OT1}{cmtt}{m}{it}{
        <->    \ttscale cmitt10
}{}
\DeclareFontShape{OT1}{cmtt}{m}{sl}{
        <->    \ttscale cmsltt10
}{}
\DeclareFontShape{OT1}{cmtt}{m}{sc}{
        <->    \ttscale cmtcsc10
}{}
\DeclareFontShape{OT1}{cmtt}{m}{ui}
       {<->ssub*cmtt/m/it}{}
\DeclareFontShape{OT1}{cmtt}{bx}{n}
       {<->ssub*cmtt/m/n}{}
\DeclareFontShape{OT1}{cmtt}{bx}{it}
       {<->ssub*cmtt/m/it}{}
\DeclareFontShape{OT1}{cmtt}{bx}{ui}
       {<->ssub*cmtt/m/it}{}

\usepackage[breaklinks=true,bookmarks=false]{hyperref}

\cvprfinalcopy 

\def\cvprPaperID{****} 

\begin{document}

\title{Sketches for Time-Dependent Machine Learning}

\author{
    Jesús Antoñanzas \\ {\tt\small jesus.maria.antonanzas@estudiantat.upc.edu}
    \and
    Marta Arias \\ {\tt\small marias@cs.upc.edu}
    \and
    Albert Bifet \\ {\tt\small abifet@waikato.ac.nz}
}

\maketitle

\begin{abstract}
Time series data can be subject to changes in the underlying process that generates them and, because of these changes, models built on old samples can become obsolete or perform poorly. In this work, we present a way to incorporate information about the current data distribution and its evolution across time into machine learning algorithms. Our solution is based on efficiently maintaining statistics, particularly the mean and the variance, of data features at different time resolutions. These data summarisations can be performed over the input attributes, in which case they can then be fed into the model as additional input features, or over latent representations learned by models, such as those of Recurrent Neural Networks. In classification tasks, the proposed techniques can significantly outperform the prediction capabilities of equivalent architectures with no feature / latent summarisations. Furthermore, these modifications do not introduce notable computational and memory overhead when properly adjusted.
\end{abstract}

\textbf{Keywords:} Machine learning, Time series analysis, Stream processing, Recurrent neural networks

\section{Introduction}
\label{sec:introduction}

Learning processes are, in many cases, highly temporal. Data is collected at specific points or periods in time and machine learning models are deployed and maintained during a set time span, all the while receiving new, more recent observations. And, as it is, the underlying, natural processes that govern these aspects, are not static. Real world dynamics affect how data is generated from one point in time to another, changing the distribution that intelligent systems had previously learned from. This phenomenon, known as 'concept drift', can potentially make valid ML models unusable. So, it is important to incorporate the ability of temporal monitorization into intelligent systems, which will provide valuable characteristics such as model autonomy and adaptation, and could make them more resilient to concept drift.

Given this strong temporal dependence of time series data and the threat of concept drift of their underlying distributions, we are interested in how much machine learning algorithms can improve their predictive performance if we incorporate relevant statistics, able to sufficiently characterize distributions, directly into the learning process.

We consider two main descriptive statistics as the core of our hypothesis: the mean and the variance. If a distribution change occurs, data previous to the drift becomes non-relevant for the current state of the model. As such, a 'forget' factor is important for the computation of relevant statistics. We have adopted the sliding window model, which only considers valid a number of the most recent data observations. In order to efficiently compute these statistics for given sliding windows, we have used sketches, which are well studied techniques from the field of data streaming to summarize information.

Firstly, we have tested how the predictive performance of traditional ML algorithms is affected when, in addition to the actual data observations, features represented by the relevant statistics of data attributes are also fed into the models. These statistics are computed on-the-fly by several Exponential Histograms (a type of sliding window sketch) of different resolutions. Then, we have tested a custom recurrent architecture with built-in summarisation capabilities. It uses Exponential Histograms (EHs) to track the changes in distribution over the hidden states, which in recurrent neural networks include important temporal information and are frequently able to represent more complex relationships than input attributes.

The contributions of this work are the following:
\begin{itemize}
    \item A proof of concept of how embedding descriptive statistics about the data distributions in learning algorithms can improve predictive performance.
    \item Experiments over classical data streaming benchmark data, exploring aspects of our proposal related to model efficiency, performance and configuration, as well as relevant comparisons.
    \item Software that includes, amongst others: (1) a Python package with several implementations of EHs able to efficiently track the mean and variance of real valued data in a user friendly framework, (2) a PyTorch implementation of our custom RNN architecture, which efficiently tracks the distribution of the data through hidden states' statistics and makes use of them when predicting, as well as its training and evaluation schedules.
\end{itemize}

In Section \ref{sec:background}, relevant background is presented. Then, in Section \ref{sec:solution} a formal and detailed description of our solution is given, with experiments in Section \ref{sec:experiments}. We use Section \ref{sec:discussion} to discuss the results of our experiments, laying out possible future directions of work in Section \ref{sec:future_work} and presenting conclusions in Section \ref{sec:conclusion}.

\section{Background}
\label{sec:background}

\subsection{Time Series}
Time series are data collections composed of points ordered in time. From acoustic signals to rain measurements, financial data to protein behaviour, query generation to network interactions or spoken languages, any measurement that is generated from a temporal process can be interpreted as a time series. Thus, time series analysis aims to extract useful statistics and relationships that can, totally or in part, explain the nature of the underlying process in order to understand it, forecast, detect anomalies or many other tasks relevant to real settings.

Common time series models are Auto-Regressive (AR) models, where the output linearly depends on previous values and a stochastic term, and Moving Average (MA) models, which model the output as a linear combination of past and current values of a stochastic term, or errors. The ideas of these two models are generalised into ARMA (auto regressive moving average), ARIMA (auto regressive integrated moving average) and ARFIMA models. When a more complex relationship between data exists, non-linear models such as Threshold AR \cite{threshold_ar}, Markov Switching \cite{markov_switching},  \cite{arch} or Hidden Markov Models \cite{HMMs} can also be used. 
Recent advances on deep learning have opened the door to more flexible time series models. To name some of the most popular ones, there are (causal) Convolutional Neural Networks, Recurrent Neural Networks such as Elman Networks \cite{ElmanNetwork} or LSTM \cite{LSTM}, or the more recent Attention-based mechanisms \cite{attention}. In spite of their flexibility, these models can be outperformed in low data regimes by more classical approaches because of their simpler generalisations \cite{time_series_DL}. Because of that, hybrid models \cite{hybrid_time_series}, which combine quantitative time series models with Deep Neural Networks, are a popular choice.

Mathematically, a time series can be considered as a set of ordered observations
\begin{equation}
    \mathcal{D} = \{\bm x_t\}_{t=1}^N, 
\end{equation}
$t = 1,...,N$ and $\bm x \in \mathbb{R}^p$. If the data points are accompanied by a target variable the set becomes 
\begin{equation}
    \label{eq:batch_time_series}
    \mathcal{D} = \{\bm x_t, y_t\}_{t=1}^N
\end{equation}
, with $y \in \mathbb{R}^1$ in univariate regression tasks or $y$ categorical in classification tasks. From here on, we consider this the problem at hand. The observations in the set $\mathcal{D}$ are assumed to be generated by processes 
\begin{equation} 
    \label{eq:static_time_series_dist}
    (\bm x_t, y_t) \sim p(\mathcal{X}, \mathcal{Y}).
\end{equation}

These processes, in most cases, generate points with some kind temporal dependence. That is, where data observations are not independent draws
\begin{equation}
    p(y_t|\bm x_t,\bm x_{t-1},\ldots,\bm x_{t-n}) \neq p(y_t|\bm x_t)
\end{equation}
for $n \in [2,t-1]$. It is this strong temporal dependence which time series models aim to exploit. 

\subsection{Data Streams}

In a traditional \textit{batch} setting, a model works with a bounded data set $\mathcal{D}$ such as in Equation \ref{eq:batch_time_series}. Once the training process is done and characteristics of the data set have been learned, the model is deployed. If new data comes that is to be learned by the model, the only alternative would be to re-train it. This, in the context of dynamic environments, is highly restrictive. Online, incremental or data stream learning aims to tackle this exact problem, adding the capacity of dynamic learning to models so that they can process data in a sequential manner, addressing the \textit{Volume} and \textit{Velocity} characteristics of big data. That is, there exist restrictions in memory and computational power when dealing with fast unbounded data streams.

In a data stream, data is assumed to arrive 
\begin{equation}
    \label{eq:dynamic_time_series_dist}
    (\bm x_t, y_t) \sim p_t(\mathcal{X}, \mathcal{Y})
\end{equation}
with $t = 1,...,\infty$.

Notice how the distribution generating the data is non-stationary (can change along $t$), in contrast with the description in Equation \ref{eq:static_time_series_dist}. When there is a change in the data distribution, then, a concept drift \cite{concept_drift} is said to have occurred. 

And, although samples from data streams are usually assumed to be i.i.d. (because change is not expected to be easily predictable), when concept drift occurs within them, a temporal dependence is exhibited \cite{CD_data_streams_are_time_series}.

Concept drift has been object of exhaustive research \cite{detection_of_changes, CD_adaptation}. The changes in the data that concept drift represents can manifest in different ways:
\begin{itemize}
    \item \textbf{Sudden}. When an abrupt change in distribution occurs, switching concepts altogether (e.g. replacement of a thermometer by a slightly different one).
    \item \textbf{Incremental}. When the change in distribution happens over time (e.g. degradation of a thermometer's capabilities over time). 
    \item \textbf{Gradual}. When the distribution switches between one concept and another with increasing frequency, finally settling in the new concept.
    \item \textbf{Reoccurring}. When the distribution switches between one concept and another periodically.
\end{itemize}

In order to deal with concept drift, several mechanisms have been studied, usually grouped into three families \cite{ML_DataStreams_MOA}. The first category encompasses methods that monitor a set of statistics, basing models on them such as Naive Bayes, which tracks frequencies of attribute and class values co-occurrences. The methods in the second family focus on detecting change and then adapting or dropping the current models, retraining them with only samples from the new concept. Lastly, model ensembles are used to dynamically represent different populations, with the capability of adapting to change.

\subsection{The Approximation Setting}

In the data streaming setting, the velocity of data samples is assumed to be too high to allow keeping them in memory. So, efficient methods and algorithms, in terms of memory and computational resources, are used. The trade-off for this efficiency, though, is not being able to provide exact answers or descriptions. Indeed, the more efficient one wants the method to be, the more relaxed will the conditions of approximation be. 

Let us formally define the approximation setting. Let $f$ be a real-valued function and let $g$ be a real-valued function defined as an approximation of $f$. Given an accuracy value $\epsilon$, $g$ is

\begin{itemize}
    \item An \textit{absolute} (additive) $\epsilon$-approximation of $f$ if 
    \begin{equation}
        \label{eq:absolute_approx}
        |f(x) - g(x)| \leq \epsilon, \forall x
    \end{equation}
    \item A \textit{relative} (multiplicative) $\epsilon$-approximation of $f$ if
    \begin{equation}
        \label{eq:relative_approx}
        |f(x) - g(x)| \leq \epsilon \cdot f(x), \forall x
    \end{equation}
\end{itemize}

The approximation function $g$ can be thought of as a random variable whose sampled values are desired to be as close as possible to the true function $f$. Another layer of flexibility can be considered in the approximation setting: an $(\epsilon, \delta)$-approximation function is defined as in Equations \ref{eq:absolute_approx} and \ref{eq:relative_approx}, but requiring that those bounds are met with probability $(1 - \delta)$. For example, a \textit{relative} $(0.01, 0.1)$-approximation of a function $f$ will return with a probability of at least $90\%$, values within $1\%$ of $f$. In a traditional, statistical sense, it is desired that the approximating function $g$ meets $\EX(g(x)) = f(x)$ and $Var(g) = \sigma^2$ where $Var(f) = \sigma^2$. 

\subsection{Sketches}

So, in the data streaming setting, the user is usually happy with an approximate answer, provided it is close enough to the ground truth and manageable amounts of resources are used. The trade-off between accuracy and memory is always present, more resources being needed if greater accuracy is required. In this context, sketches are data structures and their accompanying algorithms that are able to answer predefined queries about the stream, and are used as an alternative to regular tasks but with computational restrictions. These answers are, for the most part, given by approximation algorithms (relative or absolute, $\epsilon$ or $(\epsilon, \delta)$), fundamentally bounding the error and giving the user the possibility of adapting to different circumstances. Their aim it to use sub-linear memory and, if possible, constant computational complexity.

Popular tasks traditionally tackled by sketches are:
\begin{itemize}
    \item \textbf{Sampling}. Only a proportion of the data is sampled so that it fits into memory, and the queries are answered based on those collected data points. An example is Reservoir Sampling \cite{reservoir_sampling}, which assigns the same probability of being sampled at all times to all elements in the stream. A problem with these approaches is how they may miss infrequent, but important, items. 
    \item \textbf{Counting}. Counting the total or distinct number of elements seen, trivial when no memory or computational restriction exists, has more complex, but elegant nonetheless counterparts in the approximation setting. The Morris Counter \cite{morris_counter} approximates the total number of elements seen up to $t$ using $\log \log t$ bits. The popular HyperLogLog \cite{hyperloglog} $\epsilon$-approximates the number of distinct items seen using $O(\frac{\log \log D}{\epsilon^2})$ bits of memory, $D$ being an upper bound on the number of distinct elements.
    \item \textbf{Frequency problems}. Now, we are interested in keeping track of the number of times we have seen the most frequent distinct elements (i.e. counting the number of distinct network packages' origins and destinations seen by a router in the last 24 hours). The Space Saving sketch \cite{space_saving} can give a list of $\epsilon$-heavy hitters (frequency bigger than $\epsilon$) with memory $O\left(\frac{1}{\eps}\right)$. The Count-Min sketch \cite{cm_sketch} can solve a range of frequency problems, with the added flexibility of being able to handle both item additions and subtractions. It gives an \textit{additive} $(\eps, \delta)$-approximation of the most important (biggest) item frequencies using $O(\ln\left(\frac{1}{\delta}\right))$ memory words.
\end{itemize}

\subsection{The Sliding Window Model}

In some situations, only recent items are important for the analysis. In these cases, several strategies for constructing estimators are used, such as EWMA (Exponentially Weight Moving Average) \cite{EWMA} or the Uni Dimensional Kalman Filter (description in \cite{intro_to_kalman}). Another strategy is to adopt the \textit{sliding window} model, where only the $w$ most recent elements are considered to be relevant at all times.

Imagine that one were to use a linear sliding window to keep track of the mean of the last $w$ samples: exactly $\Theta(w)$ memory would be required. As previously mentioned, memory linearity is usually shunned in a data stream setting, as it can be prohibitive for some tasks. There are sketches that work in the sliding window domain for a variety of tasks using sub-linear memory w.r.t. $w$. One of the most prevalent in the literature is the Exponential Histogram (EH) by Datar \etal \cite{exp_hist}. Initially, it was devised as a way to continuously approximate the total number of 1's in a windowed sequence of bits. Intuitively, it works by dividing the windowed data into so called "buckets" with increasing capacity (a power of 2) the older they are. For each bucket, only the timestamp of the most recent item and its capacity (of 1's) is kept. With each new element, a sequence of merges between buckets is potentially triggered if capacities are exceeded, updating the timestamps and possibly creating new buckets. The error of the estimation comes from the last (oldest) bucket, which has an unknown number of "expired" bits. This algorithm provides a \textit{relative} $\eps$-approximation using $O(\frac{1}{\eps^2}\log^2 W)$ bits. In the same paper, a proposal for keeping the sum of a sliding window of integers was given. It works by considering each number as a sum of 1's, giving a direct translation to the original proposal.

Following papers extend the EH framework, such as the one by Babcock \etal \cite{exp_hist_var}, where a way to \textit{relatively} $\eps$-approximate the variance of real valued data points is provided with $O(\frac{1}{\eps^2} \log w)$ memory use and amortized $O(1)$ running time.

Lastly, note that the sliding window model not only is adopted for computing simple statistics, but for other tasks such as detecting change, as in the ADWIN change detector and estimator \cite{adwin1, adwin2}.


\section{Solution}
\label{sec:solution}

The premise raised in Section \ref{sec:introduction} was how / if incorporating statistics capable of characterizing the data distribution over time in the training process of machine learning models can improve their predictive performance and robustness to concept drift. With it in mind, the aim is to check if it holds in both a \textit{batch} and an \textit{online} setting. In both, the idea is to have sketches that constantly update statistics for some or all attributes of the data at hand. For each instance (at every point in time), these statistics are fed to the model as additional features. 
The range of statistics that could be kept is broad, our focus being only on the mean and the variance. Moreover, we use sliding window model because of the importance of recent data in the majority of cases. Because of this, we have adopted the EH as the sketch of choice, in particular the version from Babcock \etal \cite{exp_hist_var}, which allows to keep a \textit{relative} $\eps$-approximation of the window variance. Although the paper does not describe an accuracy bound for the window mean, the approximation that it can give has been, in our experiments, very close to that of the variance.

Multi-resolution analysis is an important tool for time series data mining, as it can discover interesting patterns that occur within different scales (an example being the Discrete Wavelet Transform \cite{DWT}).
Our approach to multi-resolution is to, instead of using one sketch of fixed size $w$, having multiple ones of different sizes (resolutions) $w_i$. This way, we can extract information at multiple scales. 

In addition to the summarisation in the input domain, we have devised a recurrent neural network architecture based on Elman Networks that uses multi-resolution statistics over the hidden states (a pooled version, really), instead of over the input features, to produce an output. The output $\tilde{y}_t$ of size $m$ of an Elman Network at time step $t$ is
\begin{equation}
    \tilde{y}_t = \sigma_y (\bm W_y h_t + b_y)
\end{equation}
where $h_t \in \mathbb{R}^h$ is the hidden state vector, $W_y \in \mathbb{R}^{m \times h}$ is a matrix of weights and $b_t \in \mathbb{R}^m$ is the bias vector. The hidden state vector $h_t$ is generated as a combination of previous hidden state vectors and the current input vector $x_t$ of size $n$ plus a bias term:
\begin{equation}
    h_t = \sigma_h (\bm W_h x_t + \bm U_h h_{t-1} + b_h)
\end{equation}
where $\bm W_h \in \mathbb{R}^{h \times n}$, $\bm U_h \in \mathbb{R}^{h \times h}$ and $b_h \in \mathbb{R}^h$ are weight matrices and vector, respectively.

\begin{figure}[h]
    \begin{center}
        \includegraphics[width = 0.9\linewidth]{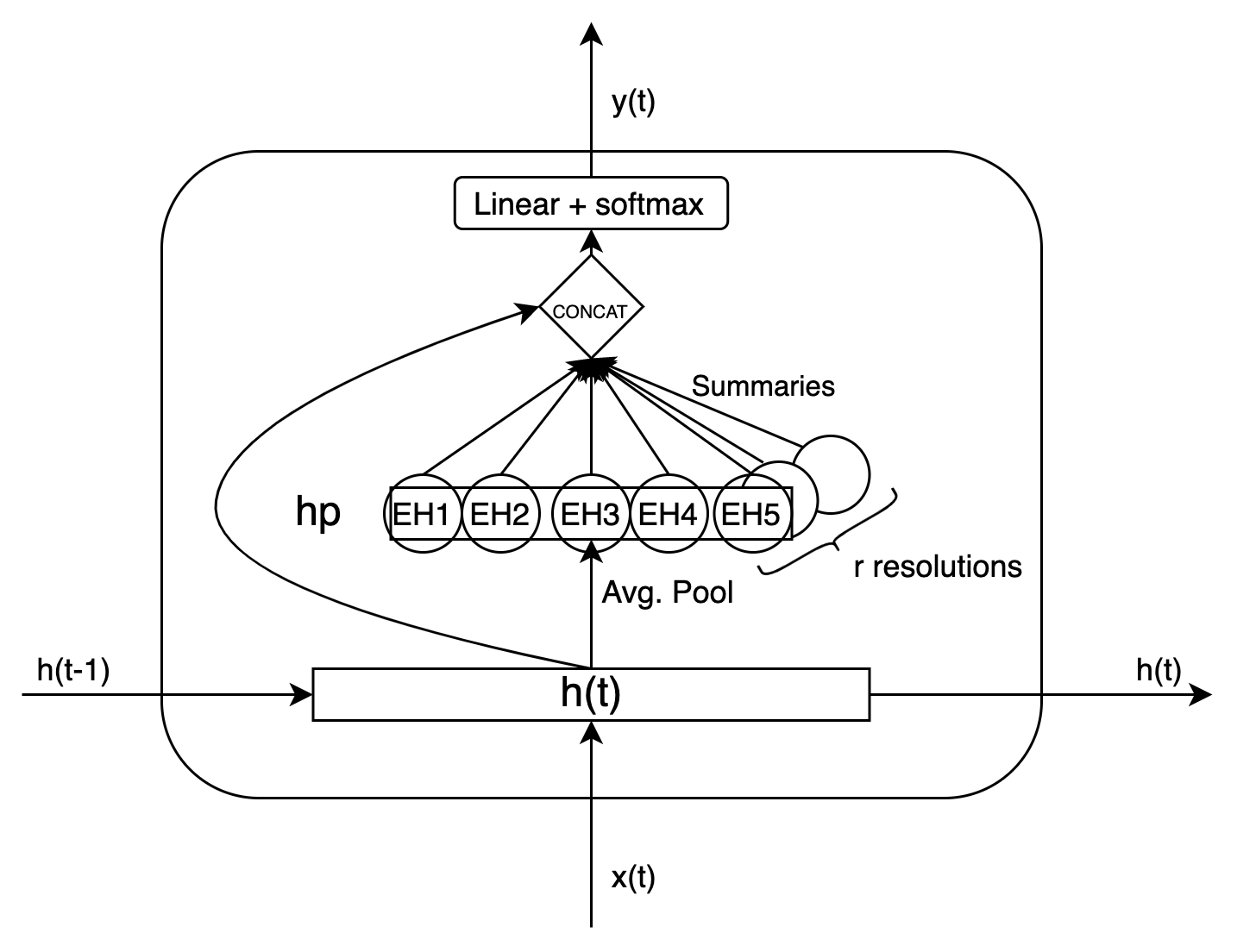}
    \end{center}
    \caption{Visual summary of a EHRNN cell for classification. At each time instant, all hidden values of the pooled hidden state vector are added to the EHs of different resolutions that track previous hidden state values. Then, statistics are estimated from these updated EHs and concatenated to the hidden state vector, the result of this being used to issue a prediction.}
    \label{fig:EHRNN_architecture}
\end{figure}

In our modified architecture, EHs keep track of the mean and variance of each element in the pooled hidden vector $h_{t}^{p}$ with different time scales. Let us define the number of elements in each $h_{t}^{p}$ as $n_p$. At each point in time, both the estimates from all features of the hidden vector and the hidden state vector itself are used to issue a new prediction. Consider $\bm E$ as a matrix of shape $n_p \times r$ that contains all of the EHs that keep track of $s$ statistics of the hidden vectors' features. That is, for each feature of a pooled hidden vector $h_{t}^{p}$, $r$ EHs are maintained, each with a different but fixed resolution. Let the evaluation $\bm E(h_{t}^{p})$ be the addition of all $n_p$ features of the pooled hidden state vector $h_{t}^{p}$ to their respective EHs (which will update their statistics) and the computation of the $n_p \cdot r \cdot s$ statistics from those EHs of different resolutions. As an example, we could have $h_t^p = 4$ hidden states, each one being monitored by $r=2$ sketches (of sizes 12 and 24, for example), which in turn output $s=1$ statistics (each one), like the mean. The prediction of the network is given by
\begin{equation}
    \tilde{y}_t = \sigma_y (\bm W_y^{\eps} h_t^{\eps} + b_y^{\eps})
\end{equation}
where $h_t^{\eps} \in \mathbb{R}^{h + n_p \cdot r \cdot s}$ is the concatenation of the hidden state vector $h_t$ and the flattened results of $\bm E(h_t^p)$ and the weight matrix and vector are now $W_y^{\eps} \in \mathbb{R}^{m \times (h + n_p \cdot r \cdot s)}$ and $b_y^{\eps} \in \mathbb{R}^{m}$. Fig. \ref{fig:EHRNN_architecture} contains a visual representation of the architecture. From now on we call this architecture EHRNN (Exponential Histogram Recurrent Neural Network) for the sake of reference.

We have based the intuition for keeping statistics over the hidden states on the their definition and how they inherently codify temporal behaviours. Indeed, they are a combination of previous hidden states, but long term temporal dependencies are lost quite fast given the vector representation of $h_t$. That is, a single hidden value is hardly able to represent long term temporal patterns. This way, statistics can be used for the model to be able to interpret multi-resolution patterns in the more meaningful (to the network) hidden representation.

\section{Experiments}
\label{sec:experiments}

We carried out some experiments to test the presented hypotheses. For the streaming setting, {\tt\small MOA} \cite{MOA} and {\tt\small Scikit-multiflow} \cite{scikit_multiflow} were used. On the other hand, we used PyTorch for experiments in the batch setting. For data preparation, all scripts were written in Python. Streaming experiments were run on a desktop machine, 8GB RAM, 2.6GHz and batch experiments on a lite Google Colaboratory machine. 

\subsection{Streaming experiments}
\label{subsec:streaming_experiments}

\begin{figure}[h]
    \begin{center}
        \subfigure[Sudden]{\includegraphics[width = 0.45\linewidth]{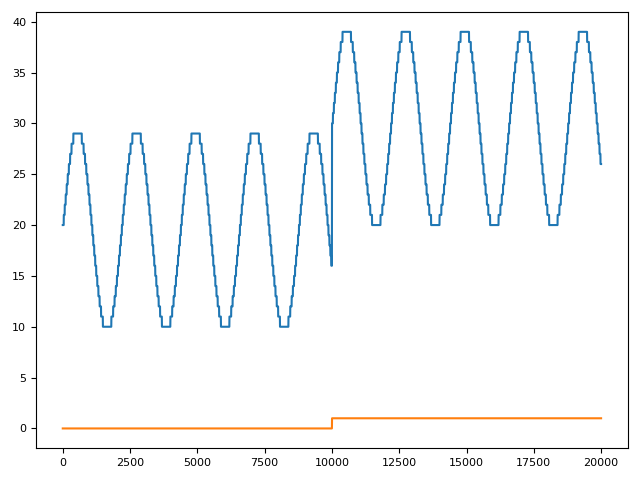}}
        \subfigure[Incremental]{\includegraphics[width = 0.45\linewidth]{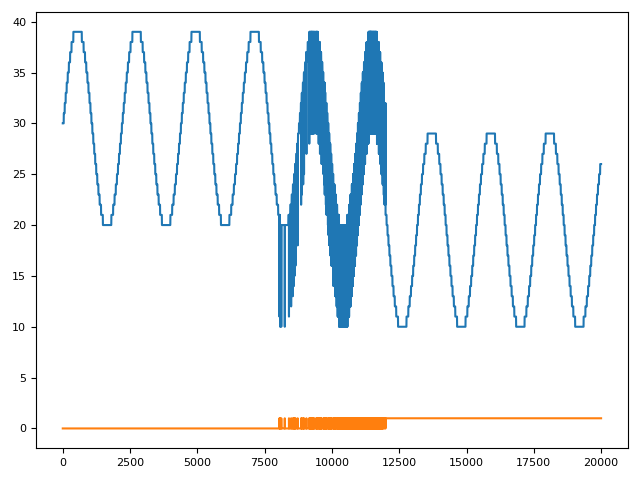}} \\
        \subfigure[Reoccurring]{\includegraphics[width = 0.45\linewidth]{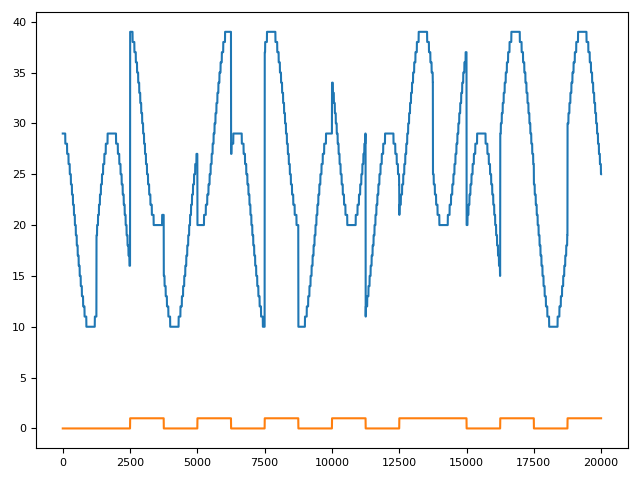}}
    \end{center}
    \caption{The different drifts generated for the synthetic data set \textbf{sineMixed}.}
    \label{fig:sineDataSets}
\end{figure}

First, synthetic data was generated. All of the synthetic data sets were based on the concatenation of sampled discrete sine waves with the shape $y_i = 10 \cdot (\sin(\theta_t) + a_i)$ at different angles $\theta_t$ and with offsets $a_i$. In particular, the data set (\textbf{sineMixed}) was sampled with sine waves parameterized at $a_1 = 2, a_2 = 3$ such that half of the range of values could overlap along the $y_i$ axis. The two concepts were represented by a each offset $a_i$ and the ground truth was to which concept each data point belonged. Over \textbf{sineMixed}, three drifts were added: \textit{sudden} (the transition between the two concepts happen at a specific time $\tau$), \textit{incremental} (points from each concept were randomly sampled following a Bernoulli distribution with increasing probability $p$) and \textit{re-occurring} (shuffled fixed-length chunks of each concept along the horizontal axis) (Fig. \ref{fig:sineDataSets}). 

\begin{singlespace}
    \begin{table}[h]
        \centering
        \tabcolsep=0.11cm
        \begin{tabular}{|c|c|c|c|c|}
            \hline
                Model & Drift & \textbf{Only win.} & \textbf{Win. + val.} & \textbf{Raw} \\
            \hline\hline
                N.B. & Sudden & 99.9 & 99.9 & 76.5 \\ 
                A.H.T. & Sudden & 99.9 & 99.9 & 99.5 \\ 
                N.B. & Incr. & 83.9 & 86.7 & 77.9 \\ 
                A.H.T. & Incr. & 96 & 95.44 & 99.4\\ 
                N.B. & Reoccur. & 73.9 & 75.7 & 70.4 \\ 
                A.H.T. & Reoccur. & 95.36 & 95.7 & 97.2 \\ 
            \hline
        \end{tabular}
        \caption{Prequential accuracy of models for \textbf{sineMixed} (Naive Bayes and Adaptive Hoeffding Tree with {\tt\small MOA} default parameters) with the different concept drifts. Bold columns represent: (\textbf{Only win.}) the data windowed with EHs of sizes \{10, 50, 100, 500, 1000\} without the original value, just the resulting means and variances. (\textbf{Win. + val.}) The same as the previous, but with the original value in a column. (\textbf{Raw}) The original value only. We see how with such simple data, only summaries can be enough for modelling. If on top of the summaries, the original values are used, the accuracy improves in almost all cases. Moreover, for simple models such as Naive Bayes, the predictive performance is better using summaries. In the case of A.H.T., we believe the added features create unnecessary complexities, thus decreasing the accuracy of the model w.r.t. using just the original attributes.}
        \label{tab:sine_results}
    \end{table}
\end{singlespace}

We also used a classical streaming benchmark data set, \textbf{Electricity} (presented in \cite{electricity_data}, normalized by A. Bifet), which contains 45,312 instances (referring to 30 min. periods) and 6 features describing an electricity market from May the \nth{7}, 1996 to December \nth{5}, 1998. The goal, then, is to predict whether the price of electricity goes {\tt\small UP} or {\tt\small DOWN}. For both \textbf{sineMixed} (Table \ref{tab:sine_results}) and \textbf{Electricity} (Table \ref{tab:electricity_online_results}) we applied several windowings to all of the features and then evaluated the results of modelling with Naive Bayes and/or Adaptive Hoeffding Trees (A.H.T.) \cite{adaptive_hoeff_tree}.

\begin{singlespace}
    \begin{table}[h]
        \centering
        \tabcolsep=0.11cm
        \begin{tabular}{|c|c|c|}
            \hline
                \textbf{Resolutions} & \textbf{Attr. values} & \textbf{\% accu.} \\
            \hline\hline
                32 & raw+mean & 86.7 \\ 
                64 & raw+mean & 85.8 \\ 
                48 & raw+mean & 86.1 \\ 
                (32, 64) & raw+mean & 86.1 \\ 
                36 & raw & 80.6 \\ 
                - & raw. & 83.4 \\ 
            \hline
        \end{tabular}
        \caption{Prequential validation of modeling different versions of \textbf{Electricity} with A.H.T, default {\tt\small MOA} parameters. \textbf{raw+mean} refers to the use of the original features and their mean for all resolutions. \textbf{raw} refers to the original data, in the case of the previous to last row letting the model look at the 32 previous samples and with no sliding window at all in the last row. We see how accuracy is higher when summarising the input attributes, more than letting the model look at previous samples (at an equivalent quantity of parameters). Moreover, more resolutions does not necessarily mean better results.}
        \label{tab:electricity_online_results}
    \end{table}
\end{singlespace}

Prequential evaluation was used in all of the experiments to measure accuracy, which consists in sequentially using each new data point to test the model and then to train it. 
\subsection{Batch experiments}

In this setting, we tested the capabilities of the EHRNN described in Section \ref{sec:solution} against the 'vanilla' (original) Elman Network. Again, we used the \textbf{Electricity} data set. For the implementation of EHRNN, we also used the Exponential Histogram from Babcock \etal \cite{exp_hist_var} to both maintain a relative $\eps$-approximation of the variance. We mainly chose this Exponential Histogram because it works with real numbers (which hidden states are). In these experiments, the accuracy was measured over the validation set, which accounts for 15\% of the total data samples after the required point in time (ordered, as is the case for the training sets, to conserve temporal dependence).  

Find configuration for the experiments in this section in Table \ref{tab:batch_experiments_config}. These values are true unless explicitly specified in figure captions. 

\begin{singlespace}
    \begin{table*}[h]
        \centering
        \tabcolsep=0.11cm
        \begin{tabular}{|c|c|c|c|c|c|c|c|c|c|}
            \hline
            Optim. & Initial l.r. & Batch Size & Hidden Size ($h$) & RNN Layers & Pool type & Pool k. size & Tr. epochs & Summaries & $\eps$\\
            \hline\hline
            RMSProp & 0.01 & 32 & 32 & 1 & avg. & $\lfloor \sqrt{h} \rfloor$ & 15 & mean \& var. &  0.05 \\ \hline
        \end{tabular}
        \caption{Network hyperparameters for both EHRNN and vanilla RNN unless explicitly specified. Note that "Pool type", "Pool k. (kernel) size", "Summary" and "$\eps$" (relative accuracy) are only valid for EHRNN. Although 15 training epochs were performed in all cases, the model that maximized validation accuracy was the one picked as to prevent overfitting.}
        \label{tab:batch_experiments_config}
    \end{table*}
\end{singlespace}

\begin{figure*}[h]
    \begin{center}
        \subfigure[By summary]{\includegraphics[width = 0.32\linewidth]{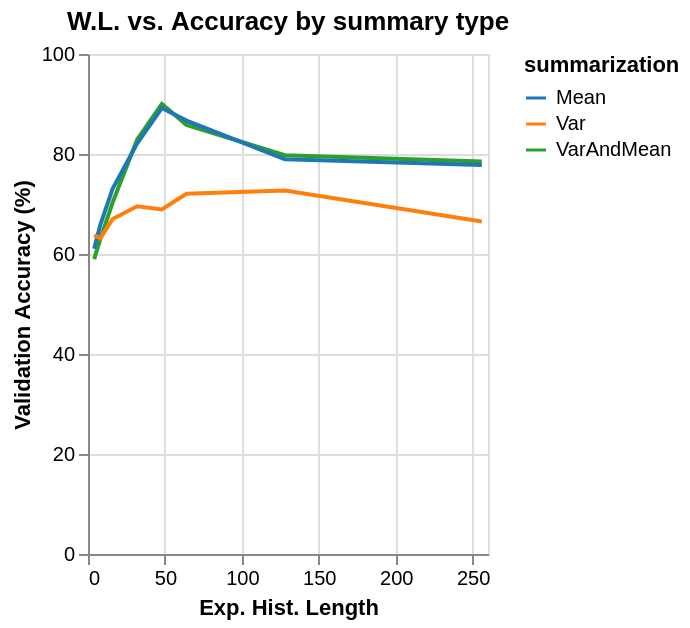}}
        \label{subfig:val_BySummary}
        \subfigure[By Hidden Size]{\includegraphics[width = 0.31\linewidth]{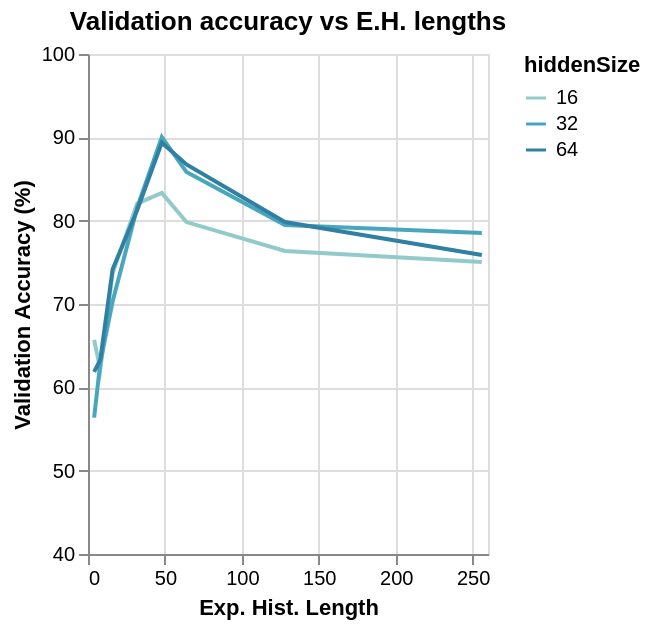}}
        \label{subfig:val_ByHiddenSize}
        \subfigure[By EH config.]{\includegraphics[width = 0.33\linewidth]{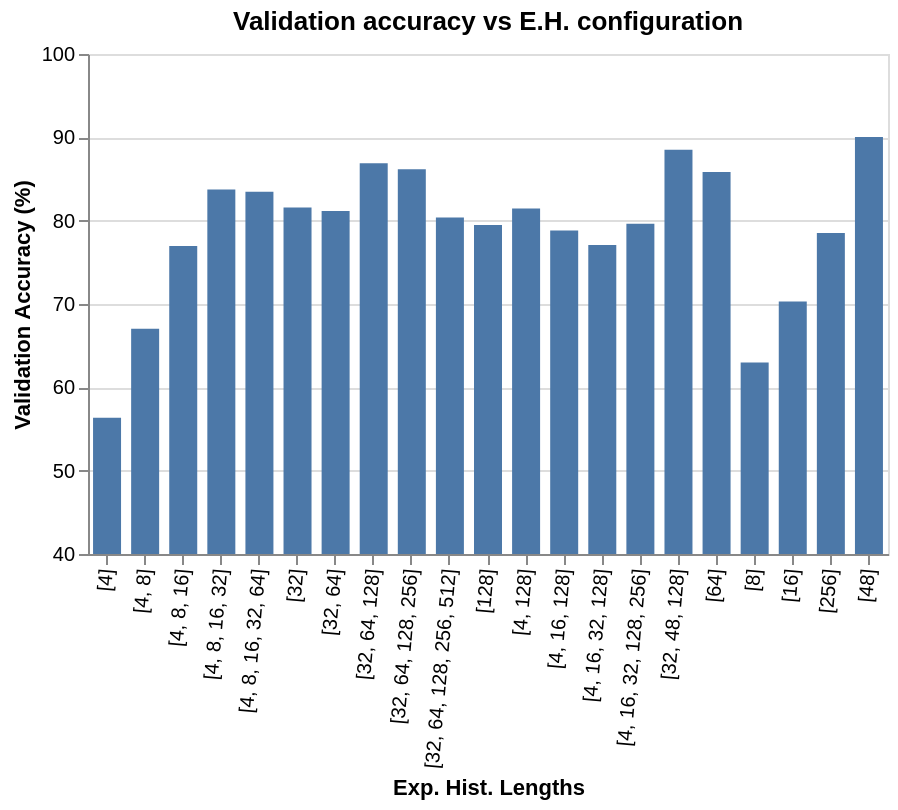}}
        \label{subfig:val_ByEHConfig}
    \end{center}
    \caption{\textbf{Electricity}. Validation accuracy of EHRNN: (a) estimating different statistics over the hidden states, (b) with different hidden sizes and (c) with different window configurations. For (a) and (b), a single EH was used over each pooled hidden state with increasing window lengths. In (b) and (c), both the mean and variance were estimated. The mean seems to be more important to the model than the variance, as the model seemingly is not affected by the latter's contribution (a). In (b), the value of hidden size is important, but a bigger one does not necessarily mean better accuracy. In both (a) and (b) we notice the importance of the window length and how a specific value seems to outperform all of the others. In (c), we note how although using different resolutions over each hidden state can drastically change validation accuracy, a window with only one resolution of value 48 outperforms all other combinations. Further discussion in Section \ref{sec:discussion}.}
    \label{fig:val_accuracies}
\end{figure*}

We tested how, given increasing window lengths, different EHRNN configurations affects the validation accuracy. In particular, we did this for summary type, for several values of hidden size and for distinct configurations of windows over the hidden states (Fig. \ref{fig:val_accuracies}).

\begin{figure}[h]
    \begin{center}
        \includegraphics[width = 0.99\linewidth]{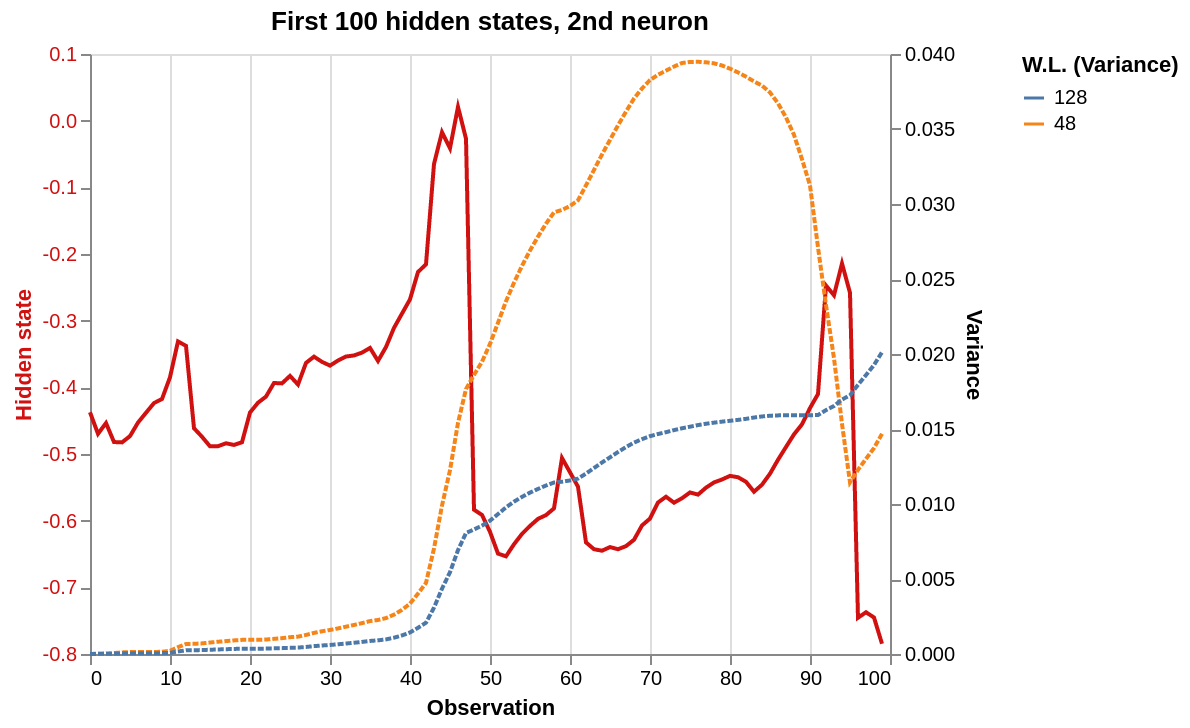}
    \end{center}
    \caption{Hidden state of a neuron EHRNN for the first 100 training points of \textbf{Electricity}. Dashed, the variance estimated with two Exponential Histograms of different resolution with $\eps = 0.05$. We notice a temporal dependence in the hidden values and a significant difference to the magnitude of the variance between the two resolutions. We notice seasonality, approximately every 48 samples. Indeed, this is the value of window size that maximizes the performance of the EHRNN over \textbf{Electricity}.}
    \label{fig:temporality_hidden_states}
\end{figure}

\textbf{Temporal dependence of hidden states}. Given how the hypotheses which we base EHRNN on is the temporal dependence of hidden values as discussed in Section \ref{sec:solution}, we wanted to visualize what was happening under the hood (Fig. \ref{fig:temporality_hidden_states}). That is, with a trained EHRNN, what the hidden values for a sequence of points were.

\begin{figure}[h]
    \begin{center}
        \subfigure[Val. accuracy]{\includegraphics[width = 0.43\linewidth]{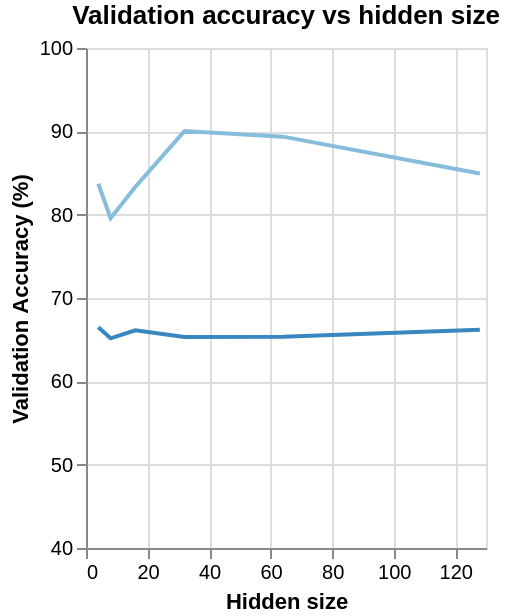}}
        \label{subfig:ehrnn_vs_rnn_Accu}
        \subfigure[Mean tr. epoch time.]{\includegraphics[width = 0.55\linewidth]{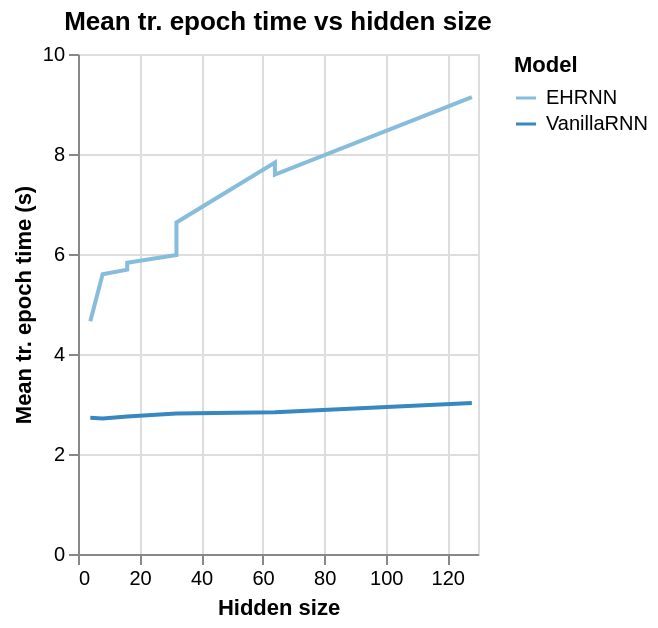}}
        \label{subfig:ehrnn_vs_rnn_EpochTime} \\
        \subfigure[Window configuration.]{\includegraphics[width = 0.99\linewidth]{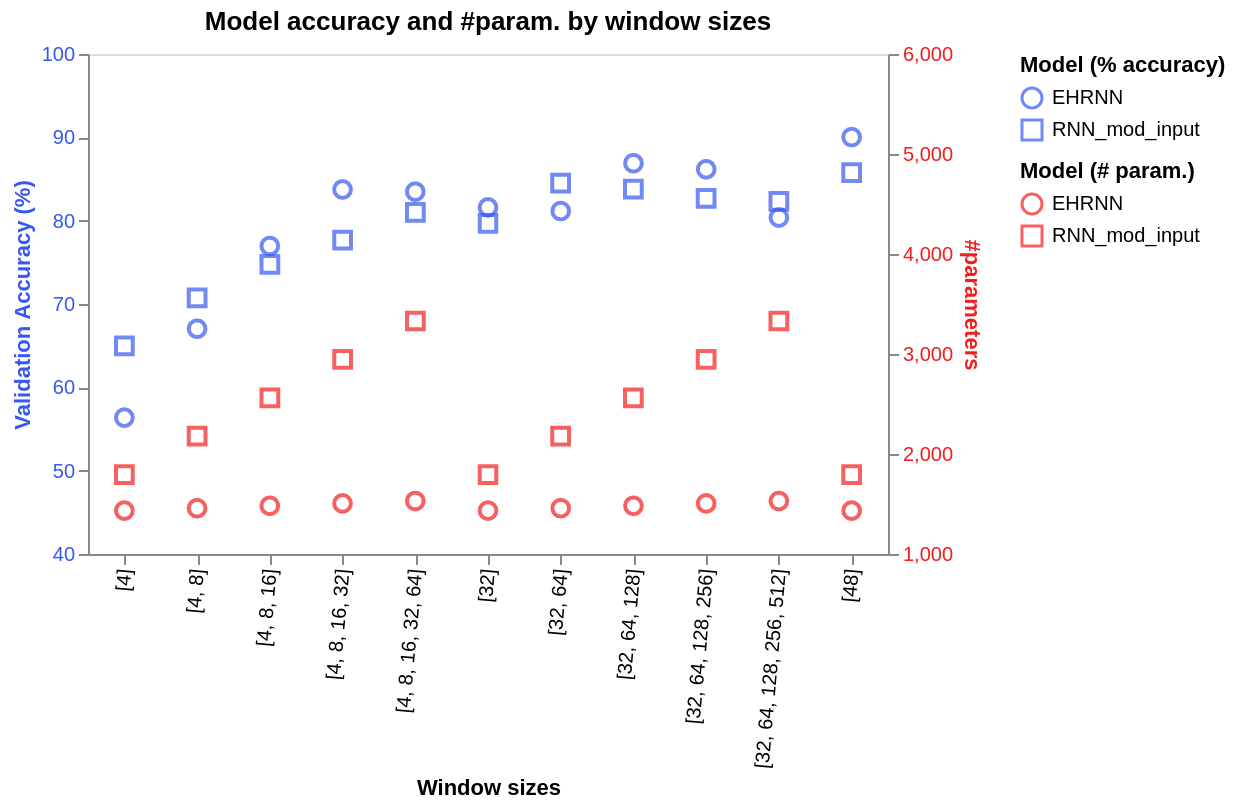}}
        \label{subfig:ehrnn_vs_rnn_WindowSizes}
    \end{center}
    \caption{\textbf{Electricity}. Comparison of: (a) validation accuracies, (b) mean training epoch time in seconds, (c) validation accuracies and number of parameters between EHRNN and the vanilla Elman Network (with windowed input, that is, concatenating mean and variance of each attribute given distinct resolutions to the data points) with same hyperparameter configuration (EHRNN window size 48). We used different hidden sizes in (a) and (b) and different window configurations for (c). We notice a consistent improvement of around $15\%$ in validation accuracy from EHRNN in comparison to not summarising over the hidden states as in vanilla RNN. In the case of (a), we see how hidden size affects in a bigger way the predictive performance of EHRNN, suggesting that too many summaries negatively the network. In (b), we see how, the bigger the size of the hidden states, the more summaries EHRNN has to keep track of, negatively impacting training time and memory usage.also increases. A nicely determined hidden size will, then, positively affect performance while not adding too much training overhead. In (c), we see how summarising over the input space (vanilla RNN) or the hidden states (EHRNN) is quite similar in terms of validation accuracy, with a slight advantage in the case of the latter. But, to achieve similar accuracies, the number of parameters in vanilla RNN is from around 50\% to roughly 700\% bigger.}
    \label{fig:ehrnn_vs_rnn}
\end{figure}

\textbf{Comparison of a vanilla RNN and EHRNN}. We compared the two architectures by parametrizing them the same way (same batch size, hidden size, layers, learning rate). The only difference between the two, then, was that EHRNN used statistics over the hidden states to compute each output. As input to the vanilla RNN, we passed both raw \textbf{Electricity} (Figures \ref{fig:ehrnn_vs_rnn} (a) and \ref{fig:ehrnn_vs_rnn} (b)) and a 'windowed' version, in a way that imitates the summarisation performed by EHRNN but over the input features like in experiments in Section \ref{subsec:streaming_experiments} (Fig. \ref{fig:ehrnn_vs_rnn} (c)). 

\section{Discussion}
\label{sec:discussion}

We discuss the results of the experiments laid out in Section \ref{sec:experiments}.

\vspace{2mm}
\textbf{Summarisation works}
\vspace{2mm} 

Using simple statistics that describe characteristics of the data at hand has improved model accuracies in the majority of our experiments. Moreover, the presented architecture EHRNN can achieve significant performance gains over the architecture it is based on with a comparable number of parameters, showing that computing the relevant statistics over hidden states is computationally more efficient and better for model accuracy than doing so in the input space. A plausible reason could be the fact that hidden states can model complex interactions or patterns in the data in a more compressed manner.

\vspace{2mm} 
\textbf{The importance of the right window resolution}
\vspace{2mm} 

Given all of the variables studied in the experiments section, the most important factor for the performance of EHRNN is window resolution. Finding the right window length to extract the maximum information possible is then a priority when trying to approach a time series modeling problem this way. For example, given how the correlation is maximum in the \textbf{Electricity} data set between samples 48 time steps apart \cite{electricity_correlation}, a sliding window of length 48 has consistently beaten other EH configurations in the experiments, as we have also seen that this correlation is brought to the hidden space (Fig. \ref{fig:temporality_hidden_states}). Then, methods for approximating this resolution on-the-fly are of interest.

\vspace{2mm} 
\textbf{Memory-time-utility tradeoff of sketches} 
\vspace{2mm} 

Although a focus has been made in using sketches as the tool for approximating statistics, their benefits are not obvious in these scales, with a caveat. That is, the benefits that can be obtained from using them usually start being relevant when there really is a restriction on either CPU or memory. When neither is present and only short temporal dependencies are present in the data, the straightforward option of computing the exact statistics can be viable. In spite of this, experiments have shown us that the relative accuracy bound $\eps$ used in EHRNN is not as important when the right scale is chosen (validation accuracy on \textbf{Electricity} hovers around 90\% with fluctuations smaller than 2\% for $\eps$ values from 0.01 to 0.6 in the case a EHRNN with configuration as in Table \ref{tab:batch_experiments_config}). Then, relaxed restrictions can be used to obtain very similar results more efficiently.

\section{Future work}
\label{sec:future_work}

In addition to using the mean and/or the variance for distribution summarisation, other statistics could be useful for different kinds of data. 
Given the importance of having the right window resolutions, we do not believe that a detached procedure for doing so is the way forward. Instead, we believe it should be possible to design an end-to-end architecture capable of adapting resolutions in the training procedure. Another solution, while not being end-to-end, could be to have ADWIN sketches, instead of fixed resolution ones.
Moreover, we believe that other common time series model architectures should be adapted to this framework. For example, statistics could be kept track of in the context of a CNN along the temporal axis. This could, for example, enhance video processing tasks such as object tracking or optical flow estimation.

\section{Conclusion}
\label{sec:conclusion}

The temporality of many learning processes is very relevant to time series analysis, as underlying data distributions can naturally change. Incorporating methods for distribution characterization in modeling tasks, then, is important for change detection and adaptability of the learning agents. With this in mind, we have tested how computing the mean and variance for specific data resolutions and feeding them into models in different manners can improve their predictive performance, using efficient methods for maintaining the statistics in sliding windows from the field of data streaming. 

While summarising over the input attributes can better the performance of models in some cases, conducting this same task over a more compressed data representation significantly improves accuracy in the tested experiments versus the same model architecture without data summaries. Not not only is it better for predictive performance, but also for efficiency. In our case, we have done so over the hidden states of an Elman Network, which represent temporal attribute information as well as attribute interactions. The tested techniques do not add significant computational overhead, but do have the restriction of having to consider different window resolutions.

We hope that other statistics or window types are tested, both static or adaptive, and that this proof of concept can be extended to more diverse tasks and model architectures in future work.

\section{Software}
\label{sec:software}
The software developed for this paper can be found in the GitHub repository \href{https://github.com/chus-chus/sketchModelling}{\textit{github.com/chus-chus/sketchModelling}}, including all of the sketch implementations, PyTorch classes (EHRNN and its associated training and validation routines) and utility functions (such as {\tt\small Pandas.DataFrame} \cite{pandas} to {\tt\small .arff} file conversion modules for quick export into machine learning frameworks such as {\tt\small MOA} \cite{MOA} or {\tt\small WEKA} \cite{WEKA}). Moreover, we have made available a Python package {\tt\small skcm} with the implemented sketches (see \href{https://test.pypi.org/project/skcm/}{\textit{PyPi Test}}), as no other Python implementations could be found. The package includes the following sliding window sketches:
\begin{itemize}
    \item Basic bit counting Exponential Histogram from \cite{exp_hist}.
    \item Integer sum Exponential Histogram with \textit{l-canonical} representation for efficient amortized update time, also from \cite{exp_hist}.
    \item Integer mean Exponential Histogram, a combination of the last two implementations. 
    \item Variance (and mean, empirically) Exponential Histogram for real valued streams from \cite{exp_hist_var}.
\end{itemize}


\newpage

{\small
\bibliographystyle{ieee_fullname}
\bibliography{bibliography}}

\end{document}